# Gradient-Guided Learning Network for Infrared Small Target Detection

Jinmiao Zhao*, Chuang Yu*, Zelin Shi†, Yunpeng Liu, Yingdi Zhang

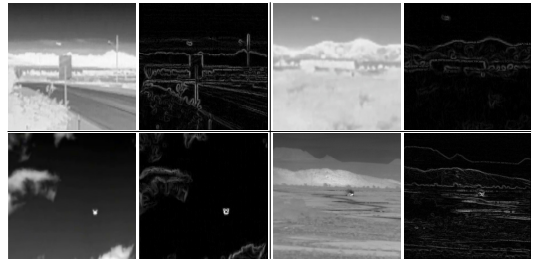

**Fig. 1.** Infrared small target image (left) and corresponding gradient magnitude image (right)

*Abstract*—Recently, infrared small target detection has attracted extensive attention. However, due to the small size and the lack of intrinsic features of infrared small targets, the existing methods generally have the problem of inaccurate edge positioning and the target is easily submerged by the background. Therefore, we propose an innovative gradient-guided learning network (GGL-Net). Specifically, we are the first to explore the introduction of gradient magnitude images into the deep learning-based infrared small target detection method, which is conducive to emphasizing the edge details and alleviating the problem of inaccurate edge positioning of small targets. On this basis, we propose a novel dual-branch feature extraction network that utilizes the proposed gradient supplementary module (GSM) to encode raw gradient information into deeper network layers and embeds attention mechanisms reasonably to enhance feature extraction ability. In addition, we construct a two-way guidance fusion module (TGFM), which fully considers the characteristics of feature maps at different levels. It can facilitate the effective fusion of multi-scale feature maps and extract richer semantic information and detailed information through reasonable two-way guidance. Extensive experiments prove that GGL-Net has achieves state-of-the-art results on the public real NUAA-SIRST dataset and the public synthetic NUDT-SIRST dataset. Our code has been integrated into https://github.com/YuChuang1205/MSDA-Net

## I. INTRODUCTION

INFRARED small target detection plays an important role in military fields such as small target tracking, infrared guidance, and airborne early warning. At present, infrared small target detection tasks face many difficulties: 1. The targets are small in size and do not have rich shape and texture features. 2. The signal-to-noise ratio of infrared images is low. 3. The scene of the infrared image is complex, and the background clutter interference is serious. 4. Lack of labeled data. Therefore, it is challenging to construct a highly accurate and robust infrared small target detection network.

Single-frame infrared small target detection is regarded as a pixel-wise segmentation task. The research can be divided into non-deep learning-based methods and deep learning-based methods. For non-deep learning-based methods, spatial filters were mainly used for detection [1] in the early days. Subsequently, non-deep learning-based methods mainly consider the high contrast between the infrared small target [2] and the background or model the infrared small target as a sparse component in a low-rank background [3], [4]. Most non-deep learning-based methods rely on static background or saliency assumptions. However, real infrared scenes are more complex and uncertain. For deep learning-based methods, Dai et al. proposed an attentional local contrast network (ALCNet) [5], which combined traditional local contrast methods with deep learning methods. Subsequently, Yu et al. successively proposed a multi-scale local contrast learning network (MLCL-Net) [6] and an attention-based local contrast learning network (ALCL-Net) [7], which further used the deep learning network to mine the local contrast information of infrared images. Recently, Li et al. proposed a densely nested attention network (DNANet) [8], which used UNet++ and a channel spatial attention module to largely preserve the information of infrared small targets in the deep network. Although the current detection method has achieved good results, it still has the problem of inaccurate edge positioning of small targets.

Limited by the characteristics of infrared long-distance imaging, the size of the target in the infrared small target image is very small, and the intrinsic features are scarce. Considering that convolutional neural networks tend to recognize targets through texture information, reasonably emphasizing edge details is helpful to improve the detection performance. From Fig. 1, the gradient magnitude image has more prominent edge textures. Therefore, we propose to introduce the gradient magnitude image into the deep learning-based infrared small target detection method, which is conducive to emphasizing the edge details and alleviating the problem of inaccurate edge positioning of small targets. At the same time, considering the importance of a reasonable feature extraction network, we construct a new dual-branch feature extraction network, which utilizes the proposed gradient supplementary module (GSM) to encode raw

This research was supported by the Innovation Project of Equipment Development Department--Information Perception Technology under Grant no. E01Z040601. *(Jinmiao Zhao and Chuang Yu contributed equally to this work.) (Corresponding author: Zelin Shi.)*

Jinmiao Zhao and Chuang Yu are with the Key Laboratory of Opto-Electronic Information Processing, Chinese Academy of Sciences, Shenyang 110016, China, also with the Shenyang Institute of Automation, Chinese Academy of Sciences, Shenyang 110016, China, also with the Institutes for Robotics and Intelligent Manufacturing, Chinese Academy of Sciences, Shenyang 110169, China, and also with the School of Computer Science and Technology, University of Chinese Academy of Sciences, Beijing 100049, China (e-mail: zhaojinmiao@sia.cn; yuchuang@sia.cn).

Zelin Shi, Yunpeng Liu and Yingdi Zhang are with Shenyang Institute of Automation, Chinese Academy of Sciences, Shenyang 110016, China (email: zlshi@sia.cn; ypliu@sia.cn; zhangyingdi@sia.cn).

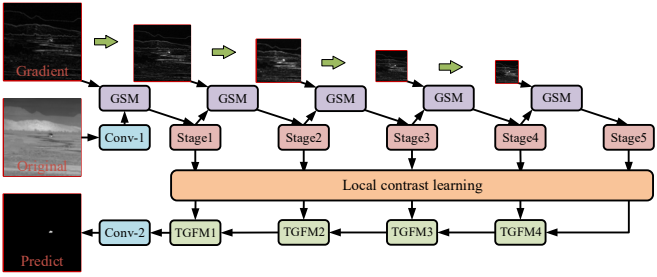

**Fig. 2.** The structure of GGL-Net. The green arrow denotes the max pooling operation.

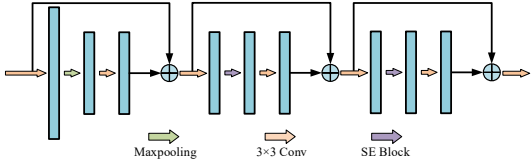

**Fig. 3.** The structure of Stage.

gradient information into deeper network layers and embeds attention mechanisms reasonably to enhance feature extraction ability. It can effectively alleviate inaccurate edge positioning and insufficient shape feature extraction in existing methods. In addition, considering that feature maps at different levels have features of different scales, we construct a two-way guidance fusion module (TGFM), which uses the spatial attention mechanism (SAM) to let the low-level features provide detailed information and spatial guidance for the high-level features, and uses the channel attention mechanism (CAM) to let the high-level features provide semantic information and channel guidance for the low-level features. It fully considers the characteristics of feature maps at different levels. In summary, we propose an innovative gradient-guided learning network (GGL-Net), which is an end-to-end network and focuses on refining infrared small target edge details. It is worth mentioning that the proposed GGL-Net achieves SOTA results on both the real NUAA-SIRST dataset and the synthetic NUDT-SIRST dataset. The contributions of this paper can be summarized as follows:

(1) We are the first to explore the introduction of gradient magnitude images into the deep learning-based infrared small target detection method, which is conducive to emphasizing the edge details and alleviating the problem of inaccurate edge positioning of small targets.

(2) We propose a novel dual-branch feature extraction network that utilizes the proposed gradient supplementary module (GSM) to encode raw gradient information into deeper network layers and embeds attention mechanisms reasonably to enhance feature extraction ability.

(3) A two-way guided feature fusion module is constructed, which fully considers the characteristics of feature maps at different levels. It can facilitate the effective fusion of multi-scale feature maps and extract richer semantic information and detailed information through reasonable two-way guidance.

## II. METHOD

### A. Gradient-guided learning network

As shown in Fig. 2, the proposed GGL-Net consists of three parts: feature extraction, local contrast learning, and feature fusion. In the feature extraction part, we propose a novel dual-branch feature extraction network to effectively improve the edge feature extraction ability of infrared small targets, which contains the main branch, supplementary branch, and connection part. In the local contrast learning part, we introduce the local contrast learning module [6], [7]. In the feature fusion part, we adopt the proposed TGFM, which makes full use of the characteristics of multi-scale features and promotes the full fusion between them.

### B. Dual-branch feature extraction network

Considering the lack of intrinsic features of infrared small targets, we propose a novel dual-branch feature extraction network that utilizes the proposed gradient supplementary module (GSM) to feed multi-scale gradient magnitude images into the main branch. By encoding the raw gradient information into deeper network levels, the network can better pay attention to the edge structure of small targets. At the same time, emphasizing the edge information of small targets in each Stage, it can greatly reduce the risk of target features being overwhelmed by background features.

As shown in Fig. 2, the main branch mainly consists of five Stage modules. The structure of Stage continues our previous work ALCL-Net [7] and makes some modifications. From Fig.3, each stage consists of six convolutional layers, which are equally divided into three blocks. At the same time, the residual connection and SE attention module [9] are used. The channel compression ratio in the SE attention module is set to 4. The SE attention module remodels the dependencies between different channels. The supplementary branch mainly consists of a set of gradient magnitude images that have been through max pooling.

For the connection part, we use the proposed GSM. From Fig. 4. GSM consists of G_Block and Res. The G_Block is used to perform feature extraction on the gradient magnitude image. The Res is similar to the block structure. Compared with simple element-wise addition, the Res structure can provide supplementary information for the main feature extraction branch more reasonably and efficiently.

### C. Two-way Guided Fusion Module

In general, the higher-level feature maps contain more semantic information but have poorer perception of details, while the lower-level feature maps contain more detailed information but insufficient understanding of scene semantics. Therefore, we propose a two-way guidance fusion module (TGFM) to promote multi-scale feature fusion. From Fig. 5, TGFM utilizes a spatial attention mechanism [10] to enable low-level features to provide detailed information and spatial guidance to high-level features. At the same time, it utilizes a channel attention mechanism to enable high-level features to provide semantic information and channel guidance to low-level features. They can be formulated as follows:

$$Z = C(Y) \otimes X + S(X) \otimes Y \tag{1}$$

$$C(Y) = \sigma(MLP(AvgPool(Y)) + MLP(MaxPool(Y))) \tag{2}$$

$$S(X) = \sigma(f^{7\times 7}([AvgPool(X); MaxPool(X)])) \tag{3}$$

where $\otimes$ denotes the element-wise multiplication. $X$, $Y$ denote the low-level feature map and the high-level feature map.

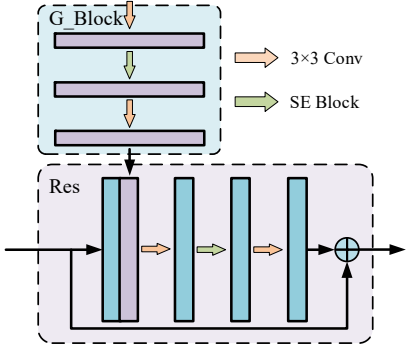

**Fig.4.** The structure of the gradient supplementary module.

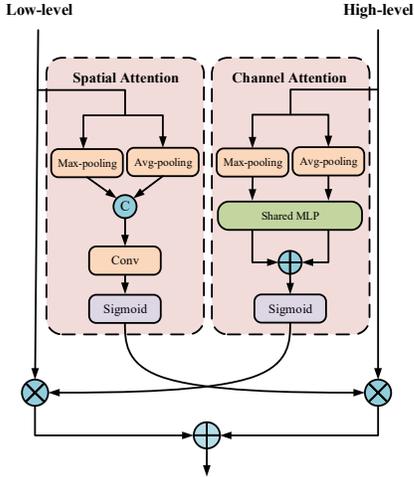

**Fig. 5.** The structure of the two-way guidance fusion module.

### D. Loss function

To address the problem of unbalanced backgrounds and targets, we use softIoU loss. It can be formulated as follows:

$$Loss_{softIoU}(p,y) = \frac{\sum_{i,j} p_{i,j} \cdot y_{i,j}}{\sum_{i,j}(p_{i,j} + y_{i,j} - p_{i,j} \cdot y_{i,j})} \quad (4)$$

where $p$ denotes the predicted score map obtained by the network, and $y$ denotes the true mask label.

## III. EXPERIMENT

### A. Dataset

*(1) NUAA-SIRST dataset:* This dataset is a real dataset [5] containing 427 images of infrared small targets from different scenes. For this dataset, 96 images are used as the test set. We uniformly resize the images to 512×512 pixels.

*(2) NUDT-SIRST dataset:* The dataset is a synthetic dataset [8] with five main background scenes: city, field, highlight section, ocean, and clouds. For this dataset, we used both 1:1 and 7:3 dataset division rules. We uniformly resize the images to 256×256 pixels.

### B. Experimental environment and parameter settings

The experimental environment is Ubuntu 18.04.6, and the GPU is an RTX 2080Ti 11G. The batch size, learning rate, and epoch are 4, 1e-4, and 500, respectively.

For evaluation metrics, we mainly use the two pixel-level evaluation metrics, the intersection-over-union (IoU) and normalized intersection-over-union (nIoU) [5]-[8]. Meanwhile, we introduce two target-level evaluation metrics, the probability of detection (Pd) and false alarm rate (Fa) [8]. In addition, we also introduce the 3D-ROC [11] curve to further prove the effectiveness of the proposed network.

### C. Ablation Study

*(1) Effect verification of the dual-branch feature extraction network:* We perform detailed comparative experiments. From Table I, "Original" and "Gradient" denote that only the original image or gradient magnitude image is input to the main branch. For each of the others, the former denotes the input of the main branch, and the latter denotes the input of the supplementary branch. First, the results of "Original+Original" do not show a significant improvement compared with the "Original" method. Instead, there is even a slight decline. A similar situation exists for the results of "gradient" and "gradient+gradient". This shows that simply adding a supplementary branch cannot have a positive effect. Second, when comparing the results of "Original+Original" and "Gradient+Gradient" with "Gradient+Original and "Original+Gradient", it can be found that combining the gradient magnitude image and the original infrared image can greatly improve the detection performance. Finally, compared to "Original", "Original+Gradient" show an increase of 1.98% (from 0.7984 to 0.8142) in IoU and 0.80% (from 0.7796 to 0.7858) in nIoU. Fig. 6 intuitively shows that compared with the "Original", the proposed GGL-Net can effectively alleviate false positives, false negatives and edge inaccuracies.

*(2) Effect verification of the gradient supplement module:* We have proposed two network variants based on GGL-Net. From Table II, "M_G_Add" denotes that the Res module is replaced by a simple element-wise addition. "M-G-M_Res" denotes that G_Block is placed between the max pooling of the gradient supplementary branch. On the one hand, replacing the Res structure with element-wise addition leads to a decrease of 1.28% (from 0.8142 to 0.8038) in IoU and 0.78% (from 0.7858 to 0.7797) in nIoU. This demonstrates that the Res structure helps to add gradient magnitude information to the main branch. On the other hand, compared with GGL-Net, directly adding G_Block to the gradient supplementary branch leads to a decrease in both IoU and nIoU. This is because the proportion of original information in the supplementary information obtained will decrease when G_Block participates in the training of the supplementary branch.

*(3) Effect verification of the two-way guidance fusion module:* For the hyperparameter $r$ in the channel attention of the TGFM, we conduct multiple experiments. From Fig. 7, we find that the value of $r$ has little impact on the final results. Meanwhile, when $r$ becomes larger, the number of parameters decreases. Therefore, we set $r$ to 8.

To fully verify the effect of the TGFM, we implement three variants on the TGFM. From Table III, "ADD" denotes direct element-wise addition, "CAM" denotes adding only channel attention from high-level to low-level, and "SAM" denotes adding only spatial attention from low-level to high-level. Compared with "ADD", "CAM" and "SAM" have better results. We find that whether low-level features provide detailed information and spatial guidance for high-level features or high-level features provide semantic information and channel guidance for low-level features can play a positive role in the network. In addition, when adding the two-way guidance, the IoU and nIoU metrics increased by 0.99% (from 0.8062 to 0.8142) and 0.77% (from 0.7798 to 0.7858), respectively. Experimental results show that making full use of the respective advantages of high-level features and low-level features can effectively improve the fusion effect of different scale features.

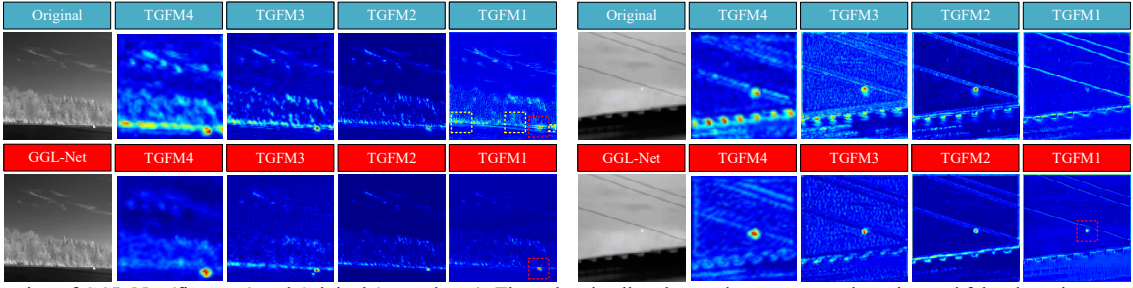

**Fig. 6.** Visualization of GGL-Net (first row) and Original (second row). The red and yellow boxes denote correct detection and false detection, respectively.

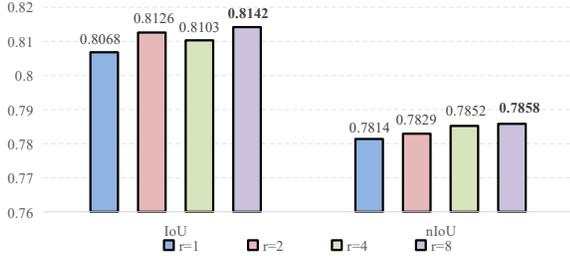

**Fig. 7.** Choice of hyperparameter r in the TGFM.

TABLE I
COMPARISON OF THE DUAL-BRANCH FEATURE EXTRACTION NETWORK AND ITS VARIANTS.

| Scheme | IoU | nIoU |
|---|---|---|
| Original | 0.7984 | 0.7796 |
| Gradient | 0.8015 | 0.7750 |
| Original+Original | 0.7889 | 0.7793 |
| Gradient+Gradient | 0.8002 | 0.7752 |
| Gradient+Original | 0.8081 | 0.7833 |
| Original+Gradient (GGL-Net) | **0.8142** | **0.7858** |

TABLE II
COMPARISON OF THE RESULTS OF THE GSM AND ITS VARIANTS.

| Scheme | IoU | nIoU |
|---|---|---|
| M_G_Add | 0.8038 | 0.7797 |
| M-G-M_Res | 0.8108 | 0.7817 |
| M_G_Res (GGL-Net) | **0.8142** | **0.7858** |

TABLE III
COMPARISON OF TGFM AND ITS VARIANTS.

| Scheme | IoU | nIoU |
|---|---|---|
| ADD | 0.8062 | 0.7798 |
| CAM | 0.8114 | 0.7810 |
| SAM | 0.8119 | 0.7812 |
| TGFM | **0.8142** | **0.7858** |

TABLE IV
COMPARISON WITH DIFFERENT METHODS ON THE NUAA-SIRST DATASET.

| Scheme | IoU | nIoU | Time on GPU/s |
|---|---|---|---|
| FKRW[1] | 0.268 | 0.339 | - |
| MPCM[2] | 0.357 | 0.445 | - |
| IPI[3] | 0.466 | 0.607 | - |
| NIPPS[4] | 0.473 | 0.602 | - |
| ALCNet[5] | 0.757 | 0.728 | 0.003 |
| MLCL-Net[6] | 0.772 | 0.755 | 0.013 |
| DNA-Net[8] | 0.784 | 0.759 | 0.111 |
| ALCL-Net[7] | 0.792 | 0.774 | 0.013 |
| **GGL-Net (Ours)** | **0.814** | **0.786** | 0.032 |

### D. Comparison of methods on the NUAA-SIRST dataset

We compare GGL-Net with eight state-of-the-art infrared small target detection methods on the NUAA-SIRST dataset. Notably, ALCNet uses the Mxnet framework, while other networks use the PyTorch framework. From Table IV, compared with ALCNet, MLCL-Net, DNA-Net and ALCL-Net, GGL-Net increases the IoU by 7.53% (from 0.757 to 0.814), 5.44% (from 0.772 to 0.814), 3.83% (from 0.784 to 0.814), and 2.78% (from 0.792 to 0.814). Moreover, GGL-Net increases the nIoU by 7.97% (from 0.728 to 0.786), 4.11% (from 0.755 to 0.786), 3.56% (from 0.759 to 0.786), and 1.55% (from 0.774 to 0.786). We also compare the single image inference time on the GPU. Compared with ALC-Net, MLCL-Net and ALCL-Net, GGL-Net significantly improves the accuracy, and the inference time increment brought by it is acceptable. Meanwhile, compared to the latest work DNA-Net, GGL-Net achieves a significant improvement in accuracy while increasing the inference speed by nearly 3.5 times. In addition, it can be found from Fig. 8 that GGL-Net has the best detection accuracy.

### E. Comparison of methods on the NUDT-SIRST dataset

We also conduct comparative experiments on the NUDT-SIRST dataset. The experimental results are shown in Table V. Bold denotes the best result, and underlined denotes the second best result. From Table V, regardless of the dataset division of 1:1 or 7:3, GGL-Net has achieved excellent results on the NUDT-SIRST dataset. Specifically, compared with ALCL-Net under the 1:1 data division, GGL-Net increased by 1.54% (from 0.909 to 0.923) and 1.41% (from 0.921 to 0.934) in IoU and nIoU, respectively. Compared with ALCL-Net under the 7:3 data division, GGL-Net increased by 1.73% (from 0.924 to 0.940) and 1.73% (from 0.924 to 0.940) in IoU and nIoU, respectively. In addition, it can be found from Fig. 9 that GGL-Net has the best infrared small target detection and background suppression effect.

## IV. CONCLUSION

This paper proposes an innovative end-to-end gradient-guided learning network (GGL-Net) for single-frame infrared

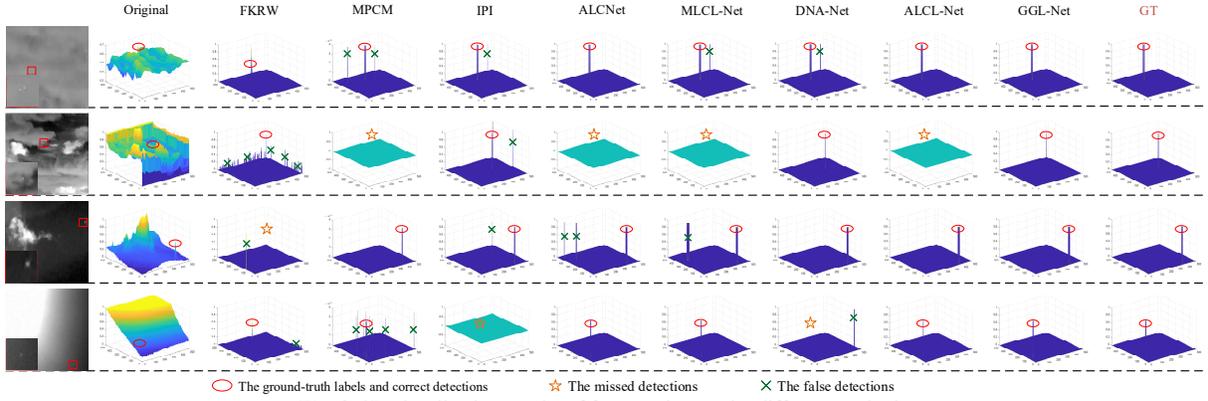

Fig. 8. 3D visualization results of four test images by different methods.

○ The ground-truth labels and correct detections    ☆ The missed detections    × The false detections

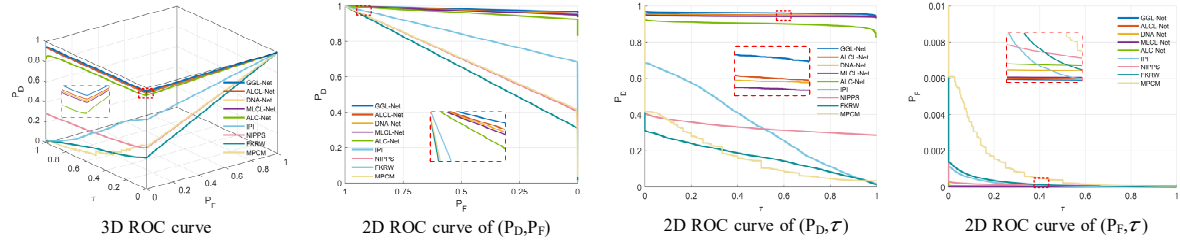

3D ROC curve    2D ROC curve of ($P_D$, $P_F$)    2D ROC curve of ($P_D$, $\tau$)    2D ROC curve of ($P_F$, $\tau$)

**Fig.9.** 3D ROC curves along with their generated three 2D ROC curves using the NUDT-SIRST dataset (7:3) and uniform step size, $\triangle = 0.0001$.

TABLE V
COMPARISON WITH DIFFERENT METHODS ON THE NUDT-SIRST DATASET.

| Methods | IoU | | nIoU | | Pd | | Fa (×10⁻⁶) | | Time on GPU/s |
|---|---|---|---|---|---|---|---|---|---|
| | 1:1 | 7:3 | 1:1 | 7:3 | 1:1 | 7:3 | 1:1 | 7:3 | |
| FKRW[1] | 0.110 | 0.116 | 0.232 | 0.241 | - | - | - | - | - |
| MPCM[2] | 0.123 | 0.110 | 0.198 | 0.147 | - | - | - | - | - |
| IPI[3] | 0.403 | 0.352 | 0.497 | 0.457 | - | - | - | - | - |
| NIPPS[4] | 0.279 | 0.315 | 0.180 | 0.365 | - | - | - | - | - |
| ALCNet[5] | 0.822 | 0.830 | 0.835 | 0.844 | **0.990** | 0.979 | 7.24 | 13.07 | 0.003 |
| DNA-Net[8] | 0.873 | 0.883 | 0.879 | 0.884 | 0.989 | 0.991 | **3.01** | 2.59 | 0.028 |
| MLCL-Net[6] | 0.895 | 0.912 | 0.904 | 0.914 | 0.970 | **0.993** | 7.84 | 15.36 | 0.005 |
| ALCL-Net[7] | 0.909 | 0.924 | 0.921 | 0.924 | 0.986 | 0.990 | 7.47 | 3.61 | 0.008 |
| **GGL-Net (Ours)** | **0.923** | **0.940** | **0.934** | **0.940** | 0.989 | **0.993** | 4.44 | **2.39** | 0.016 |

small target detection. On the one hand, considering the lack of intrinsic features and blurred edges of infrared small targets, we propose a novel dual-branch feature extraction network that can strengthen the attention to the edge details of the target. On the other hand, considering the problem of semantic and scale discontinuity of multi-scale features, we propose a two-way guidance fusion module, which fully considers the characteristics of feature maps at different levels and achieves sufficient fusion of features at different scales. In addition, extensive experiments show that GGL-Net achieves state-of-the-art results on both the real NUAA-SIRST dataset and the synthetic NUDT-SIRST dataset.